\documentclass{article}

\usepackage[final]{corl_2025} 

\usepackage{microtype}
\usepackage{graphicx}
\usepackage{subfigure}
\usepackage{booktabs} 
\usepackage{setspace}
\usepackage{hyperref}
\usepackage{url}
\usepackage{amsmath, amssymb, amsfonts}
\usepackage{algorithm}
\usepackage{algpseudocode}
\usepackage{array}
\usepackage{bbm}
\usepackage[frozencache,cachedir=.]{minted}
\usepackage[table]{xcolor}

\definecolor{LightGray}{gray}{0.97}

\newcommand{\Ours}{POMDP Coder }
\newcommand{\OursNoSpace}{POMDP Coder}

\title{LLM-Guided Probabilistic Program Induction for POMDP Model Estimation}

\author{
   \textbf{Aidan Curtis}$^{1}$, \textbf{Hao Tang}$^{2}$, \textbf{Thiago Veloso}$^{1}$,
   \textbf{Kevin Ellis}$^{2}$, \\\textbf{Joshua Tenenbaum}$^{1}$, \textbf{Tomás Lozano-Pérez}$^{1}$, \textbf{Leslie Pack Kaelbling}$^{1}$ \\\\
   $^1$MIT \quad $^2$Cornell University
}

\begin{document}
\maketitle


\begin{abstract}
    Partially Observable Markov Decision Processes (POMDPs) model decision making under uncertainty. While there are many approaches to approximately solving POMDPs, we aim to address the problem of learning such models. In particular, we are interested in a subclass of POMDPs wherein the components of the model, including the observation function, reward function, transition function, and initial state distribution function, can be modeled as low-complexity probabilistic graphical models in the form of a short probabilistic program. Our strategy to learn these programs uses an LLM as a prior, generating candidate probabilistic programs that are then tested against the empirical distribution and adjusted through feedback. We experiment on a number of classical toy POMDP problems, simulated MiniGrid domains, and two real mobile-base robotics search domains involving partial observability. Our results show that using an LLM to guide in the construction of a low-complexity POMDP model can be more effective than tabular POMDP learning, behavior cloning, or direct LLM planning. 
\end{abstract}



\section{Introduction}
Decision making under uncertainty is a central challenge in robotics, autonomous systems, and artificial intelligence more broadly. Partially Observable Markov Decision Processes (POMDPs) provide a principled framework for modeling and solving such problems by explicitly representing uncertainty in state perception, transitions, and rewards. Many prior works have used the POMDP formulation to solve real-world problems such as intention-aware decision-making for autonomous vehicles~\cite{autonomous_driving}, collaborative control of smart assistive wheelchairs~\cite{wheelchair}, robotic manipulation in cluttered environments~\cite{grasping_pomdp}, and generalized object search~\cite{generalized_object_search}.
Despite their conceptual clarity, practical application of POMDPs is bottlenecked by difficulties in specifying accurate models of environments, which requires careful engineering and a thorough understanding of the theory and available solvers.

In this work, we address the critical challenge of learning interpretable, low-complexity POMDP models directly from data. We specifically target a class of POMDPs whose components, namely the observation function, reward function, transition dynamics, and initial state distribution, can be succinctly represented as probabilistic graphical models encoded by short probabilistic programs. To efficiently identify these programs, we leverage recent advancements in Large Language Models (LLMs) to serve as informative priors, generating candidate probabilistic programs. These candidates are evaluated against empirical observations and iteratively refined through LLM-generated feedback.

We evaluate our approach across several domains including classical POMDP problems, minigrid navigation and manipulation problems, and a real-world robotics setting involving a mobile-base robot searching for a target object. Our experimental results demonstrate that guiding model construction with an LLM significantly enhances sample efficiency compared to traditional tabular model or behavior learning methods and achieves greater accuracy than directly querying an LLM.

\section{Related Work}
\label{sec:related_work}

Learning world models for partially observable decision-making is a form of model-based reinforcement learning ~\cite{mbrl}, for which there are many methods. In this section, we focus on methods that emphasize extreme data efficiency through the use of explicit models and representations, such as probabilistic programs, that facilitate efficient learning. Additionally, we discuss methods that use LLMs to synthesize and refine these models, enabling the integration of human priors and domain-specific constraints to create world models for downstream solvers. 

\noindent
\textbf{POMDP Model Learning} A substantial body of research has addressed the challenge of learning Partially Observable Markov Decision Process (POMDP) models from experience. For example, Mossel and Roch ~\cite{learning_pomdps_is_hard} provide an average-case complexity result showing that estimating the parameters of certain Hidden Markov Models, an essential subproblem in POMDP learning, is computationally intractable in general. Despite these challenges, several approaches have made significant progress by introducing tractability under specific assumptions.

Bayesian methods, such as Bayes-Adaptive POMDPs ~\cite{bayes_pomdp}, incorporate model uncertainty directly into decision-making by unifying model learning, information gathering, and exploitation. This, however, increases the overall complexity of the POMDP. In contrast, spectral techniques like Predictive State Representations (PSRs) ~\cite{PSR} offer computational efficiency by bypassing full Bayesian inference, although they require strong structural assumptions and extensive exploratory data.

Recent work on optimism-based exploration algorithms has yielded theoretical guarantees for efficient learning in specific POMDP subclasses, even though these methods can be challenging to apply directly to real-world, complex domains ~\cite{undercomplete_pomdp}. Additionally, apprenticeship learning approaches estimate POMDP parameters by leveraging expert demonstrations, assuming that expert behavior encapsulates informative state-transition dynamics ~\cite{makino2012apprenticeshiplearningmodelparameters}. Such methods help reduce the burden of exploration but are sensitive to the quality of the expert demonstrations. In dialogue systems, for instance, these techniques have successfully learned user models without relying on manual annotations ~\cite{dialogue_pomdp}.

\noindent
\textbf{Probabilistic Program Induction.} Probabilistic programming provides a powerful framework for modeling complex systems using concise, symbolic representations ~\cite{pyro, gen, church}. Several works in probabilistic program induction have demonstrated that such representations can lead to markedly improved data efficiency and enable few-shot learning, in stark contrast to more data-intensive conventional methods ~\cite{dreamcoder, lake2015human}. Nonetheless, the vast search space inherent to probabilistic programs can be a major computational bottleneck. Recent advances have attempted to address this issue by integrating language models with probabilistic programming, thereby infusing human priors into the model discovery process ~\cite{li2024automatedstatisticalmodeldiscovery, wong2023wordmodelsworldmodels, grand2024looselipssinkships}.

\noindent
\textbf{LLM Model Learning for Decision-Making.} The application of large language models (LLMs) to world-modeling for decision-making is an emerging and rapidly evolving area. Prior work has primarily focused on fully observable settings, where transition dynamics and reward structures are represented using frameworks such as PDDL or code-based models ~\cite{liang2025visualpredicatorlearningabstractworld, tang2024worldcodermodelbasedllmagent}. Other approaches have utilized code-based representations as constraints or as part of optimization frameworks ~\cite{curtis2024trustproc3ssolvinglonghorizon, hao2024large, ye2023satlm}. To our knowledge, our work is the first to extend these techniques to the POMDP setting, thereby addressing the additional complexities introduced by partial observability.
\section{Background}
\label{sec:background}

\subsection{Partially Observable Markov Decision Processes}
\label{sec:background_pomdp}
Partially Observable Markov Decision Processes (POMDPs) provide a principled framework for sequential decision making under uncertainty. A POMDP is defined by the tuple \((\mathcal{S}, \mathcal{A}, \mathcal{O}, \mathcal{T}, \mathcal{Z}, \mathcal{R}, \gamma)\), where \(\mathcal{S}\), \(\mathcal{A}\), and \(\mathcal{O}\) denote the state, action, and observation spaces, respectively. 
In this work we assume all these spaces are discrete, but the POMDP formulation supports continuous spaces in general.
\(\mathcal{T}(s_{t+1}\mid s_t, a_t)\) indicates the distribution over next states expected when taking action $a_t$ in state $s_t$. 
The observation model \(\mathcal{Z}(o_t\mid s_{t+1}, a_t)\) represents the distribution over observations expected when executing action $a_t$ resulting in a subsequent state $s_{t+1}$.
\(\mathcal{R}(s_t,a_t,s_{t+1})\) is the reward function.
Lastly, \(\gamma \in (0,1)\) is the discount factor which weighs the value of current rewards over future ones.

Since the agent does not have direct access to the true state \(s_t\), it maintains a belief \(b_t(s)\), or a probability distribution over possible states, which must be updated for replanning after each action is taken and new observation received. 
Given an action and observation, a belief can be updated via \emph{particle filtering}~\cite{probabilistic_robotics}. 
We use particle filtering as our belief-updating mechanism across all domains.

\begin{figure*}[t!]
    \centering
    \includegraphics[width=0.9\textwidth]{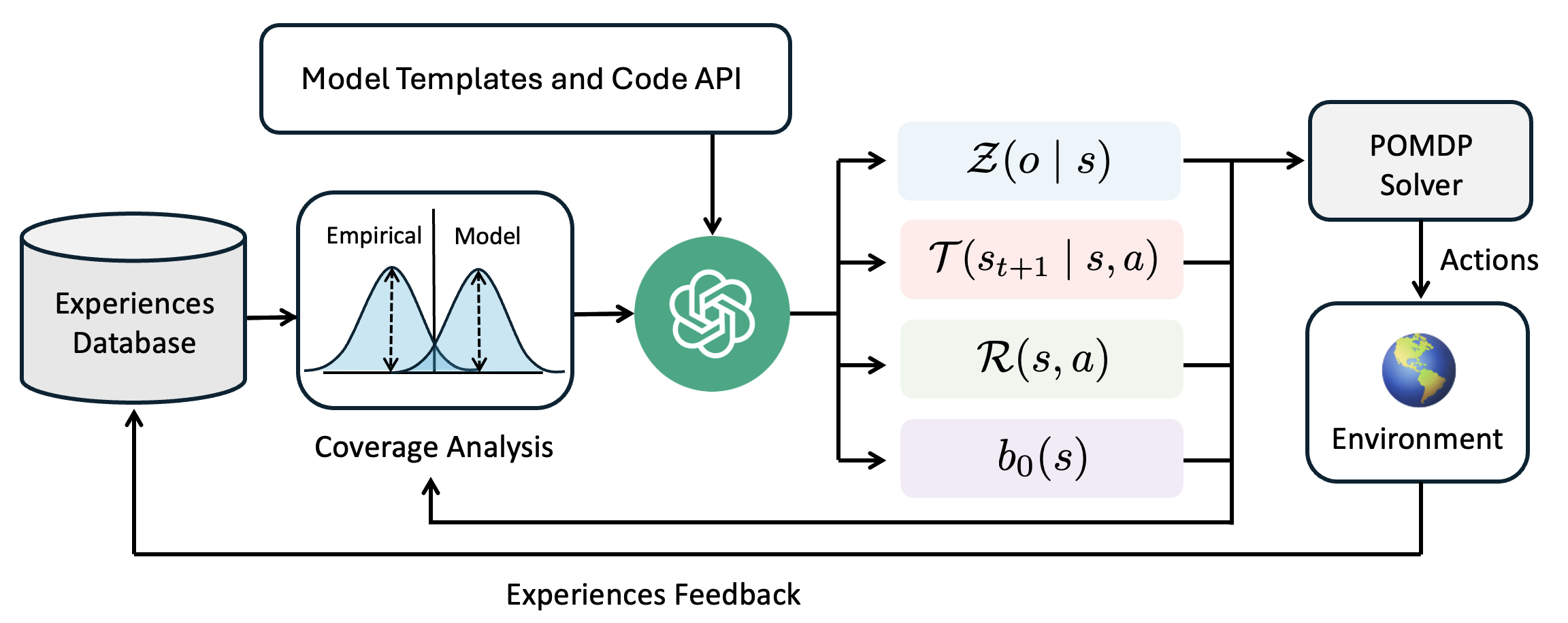}
    \caption{An architecture diagram for our POMDP coder method}
    \label{fig:arch}
\end{figure*}

The objective of a POMDP is to find a policy $\pi$ that maximizes the expected discounted reward:

\[
\max_{\pi} \; \mathbb{E}_{b_0, \pi} \left[ \sum_{t=0}^{\infty} \gamma^t \sum_{s \in \mathcal{S}} b_t(s) \sum_{s' \in \mathcal{S}} \mathcal{T}(s' \mid s, a_t) \, \mathcal{R}(s, a_t, s') \right]
\]

To illustrate the POMDP formulation and serve as a running example, consider the classic Tiger problem~\cite{tiger}. An agent faces two doors: one hides a tiger, the other a treasure. The true state \(s\) consisting of the the tiger’s location is hidden. The agent can \texttt{listen} to receive a noisy observation \(o\) of which door the tiger is behind, or \texttt{open} a door to gain a reward or incur a penalty. An ideal policy is to wait long enough to be confident in the tiger's location before opening the door with the treasure.

\subsection{Probabilistic Programs}
\label{sec:background_probabilistic_programs}
Probabilistic programming offers an expressive and concise way to represent complex probabilistic models as executable code. In our work, each component of the POMDP including the initial state model, transition dynamics, observation function, and reward structure is encoded as a short probabilistic program. We leverage \textit{Pyro}~\cite{pyro}, a flexible probabilistic programming framework built on Python, to specify these models. Pyro enables us to define generative models with inherent stochastic behavior. 
In our tiger problem example, the below implementation would be a correct probabilistic program for the observation model. 
\begin{minted}[fontsize=\small, breaklines, tabsize=4, bgcolor=LightGray]{python}

def tiger_observation_func(state: TigerState, act: TigerActions):
    if act != TigerActions.LISTEN:
        return TigerObservation(NONE)

    correct = bool(pyro.sample("listen_correct", Bernoulli(torch.tensor(0.85))))
    tiger_left = state.tiger_location == 0
    hear_left  = (correct and tiger_left) or (not correct and not tiger_left)
    return TigerObservation(HEAR_LEFT if hear_left else HEAR_RIGHT)
    
\end{minted}

\subsection{POMDP Solvers}
\label{sec:background_solvers}
While traditional offline solvers aim to compute a global policy over the entire belief space~\cite{smallwood1973optimal, littman1995witness, cassandra1997incremental}, these methods often face scalability challenges in high-dimensional or continuous environments. 
In contrast, online solvers focus on finding a solution from a specific initial belief and replan after every step as new observations become available. 
Online approaches such as Partially Observable Monte Carlo Planning (POMCP)~\cite{pomcp, curtis2022taskdirectedexplorationcontinuouspomdps} and Partially Observable Upper Confidence Trees (POUCT)~\cite{POMCPOW} combine Monte Carlo sampling with tree search to approximate optimal policies in real time.
Additionally, determinized belief space planners simplify the stochastic nature of the problem by converting it into a deterministic surrogate, thereby enabling rapid replanning~\cite{ffreplan, BHPN, tampura}. 
In our work, we adopt the determinized belief space planning approach due to its computational efficiency in larger problems. Please refer to Appendix~\ref{app:planner} for specifics on our solver implementation.

\section{POMDP Coder}
\label{sec:method}

In our approach, which we call \OursNoSpace, we decompose the problem into two major components: learning the probabilistic models that define the POMDP and using these learned models for online planning. The first part involves leveraging a Large Language Model (LLM) to generate, refine, and validate candidate probabilistic programs that represent the initial state, transition, observation, and reward functions. The second component uses these models within an online POMDP solver along with the current belief to find optimal actions to take in the environment.

We assume the agent is provided access to an initial set of ten human-generated demonstrations $\mathcal{D}$, where each demonstration consists of $(s_t, a_t, o_t, s_{t+1})$ transitions. Since we are learning models instead of policies, there are no assumptions made about the quality of these demonstrations. However, we do make an assumption of post-hoc full observability~\cite{pinto2017asymmetric}.
That is, we assume the agent gets access to the intermediate states after an episode has terminated (see Section~\ref{sec:limitations} for details).

Additionally \Ours is provided a code-based API defining the structure of the state, action, and observation space. Below is an example for the Tiger domain.

\begin{minted}[fontsize=\small, breaklines, tabsize=4, bgcolor=LightGray]{python}

class TigerActions(enum.IntEnum):
    OPEN_LEFT = 0, OPEN_RIGHT = 1, LISTEN = 2

class TigerObservation(Observation):
    obs: int # 0 = hear left, 1 = hear right, 2 = none

class TigerState(State):
    tiger_location: int  # 0 = left, 1 = right
    
\end{minted}

\begin{algorithm}[h]
\begin{spacing}{1.2}
\begin{algorithmic}[1]
\State \textbf{Input:} A demo dataset $\mathcal{D}$, max episodes $E$, num particles $N$, empty models $\theta=\emptyset$
\For{$\text{episode}=1$ \textbf{to} $E$}
    \State $\theta = (\theta_{\text{trans}}, \theta_{\text{rew}}, \theta_{\text{obs}}, \theta_{\text{init}}) \gets \texttt{LearnModels}(\mathcal{D}, \theta)$ \Comment{Update all models using $\mathcal{D}$} \label{line:learn_models}
    \State $b \gets N\text{ samples from } \theta_{\text{init}}$
    \While{episode not terminated}
         \State $a \gets \texttt{POMDPSolver}(b, \theta)$ \Comment{Plan best next action, see Appendix~\ref{app:planner}} \label{line:pomdp_solver}
         \State Execute $a$ in the world, observe $o$ \label{line:execute}
         \State $b \gets \texttt{ParticleFilter}(b, a, o, \theta)$ \Comment{Update belief} \label{line:update}
         \State Append $(a, o)$ to trajectory $\tau$ \label{line:record}
    \EndWhile
    \State $\mathcal{D} \gets \mathcal{D} \cup \tau$ \Comment{Update demo data with new trajectory}
\EndFor
\State \Return $\theta$
\end{algorithmic}
\end{spacing}
\caption{POMDP Coder}
\label{alg:pomdpcoder}
\end{algorithm}

Given these inputs, \Ours proceeds as outlined in Algorithm~\ref{alg:pomdpcoder}. It proposes an initial set of models that comprise the POMDP problem (Line~\ref{line:learn_models}), using a learning procedure detailed in Section~\ref{sec:method_learning}. After an initial set of models is decided on, these models are passed to a POMDP solver along with the initial belief (Line~\ref{line:pomdp_solver}) to find an optimal first action to take in the environment (Line~\ref{line:execute}). 
After an action is taken and an observation received, the belief is updated using particle filtering to form a new belief (Line~\ref{line:update}). This process continues until the episode terminates or times out. Lastly, at the end of each episode, the trajectory is added to the dataset (Line \ref{line:update}) and the learning process repairs any inaccuracies that the previous model may have had under the new data (Line \ref{line:learn_models}).


\subsection{Learning Models}
\label{sec:method_learning}

We aim to learn four core components of a POMDP: the initial state distribution \(P(s_0)\), the transition model \(P(s_{t+1} \mid s_t, a_t)\), the observation model \(P(o_t \mid s_{t+1}, a_{t})\), and the reward model \(R(s_t, a_t, s_{t+1})\). 
Each of these components is expressed as a short probabilistic program.

The objective of our learning procedure is to maximize a dataset $\textit{coverage}$ metric, which we define to be the proportion of data in $\mathcal{D}$ that has support under the model as follows:

\[
\text{coverage}(P_{\theta}, \mathcal{D})
\;=\;
\frac{1}{|\mathcal{D}|}\sum_{i=1}^{|\mathcal{D}|}
\mathbf{1}\!\Bigl[\,P_{\theta}\bigl(y_i \mid x_i\bigr) > 0\,\Bigr].
\]

Although we experimented with other metrics such distributional distance metrics, we found that those were more prone to overfitting and less interpretable to an LLM than binary coverage feedback. Still, the coverage metric has its own limitations, which we discuss in Section~\ref{sec:limitations}.

Our approach to learning these models builds on strategies previously developed for reward model learning~\cite{rex}, but extends them to the more general setting of POMDP model learning across multiple stochastic components (transition, observation, initial state, and reward), replaces accuracy with coverage, and introduces the notion of a testing and training split to avoid overfitting. 

At its core, our model learning strategy uses two operations: (1) \textit{LLM program proposal} given a model function template and a set of examples from the database and and (2) \textit{LLM program repair} given a previous model and set of examples that the previous model failed to cover. We run a stochastic procedure for sampling which program to repair next, which is biased toward repairing programs that have high coverage. The pseudocode and additional details can be found in the Appendix~\ref{app:model_learning}.


\section{Experiments}
\label{sec:results}

\subsection{Simulated Experiments}
\label{sec:simulated_experiments}

Simulated experiments were conducted on two categories of problems: classical POMDP problems from the literature and MiniGrid tasks. The classical POMDP problems serve as simplified benchmarks that capture the core challenges of decision making under uncertainty while keeping the problem domains small and tractable. In addition to the tiger problem, we evaluate on the rock sample problem~\cite{rock_sample}.

\begin{figure*}[ht!]
    \centering
    \includegraphics[width=1.0\textwidth]{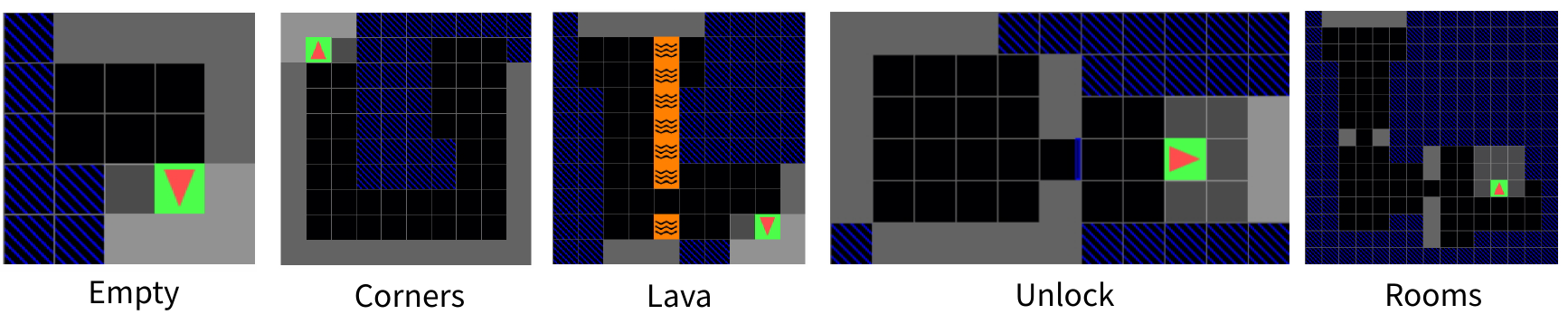}
    \caption{A visualization of the final belief state for each of the MiniGrid tasks. The green square is the goal, the red triangle is the agent, and the blue squares are places that the agent has not viewed.}
    \label{fig:envs}
\end{figure*}

MiniGrid is a set of minimalistic gridworld tasks for testing navigation and planning under partial observability originally designed for reinforcement learning~\cite{minigrid}. We evaluate on five of these environments. In the \textit{empty} environment, the agent is deterministically placed in the top left cell of a 5×5 grid and must navigate to the green square in the bottom right. In the \textit{corners} environment, the agent is randomly positioned with an arbitrary orientation in a 10×10 grid while the green square appears in a randomly selected corner. In the \textit{lava} environment, the agent starts in the upper left corner of a 10×10 grid that features a randomly positioned column of lava with a gap forming a narrow passage to the green square. In the \textit{unlock} environment, the agent is randomly placed in the left room of a two-room layout, must collect a randomly placed key to open a locked door, and then proceeds to the green square fixed at the center of the right room. In the \textit{rooms} environment, the agent is initialized in the upper right-hand corner of a multi-room setting and must traverse through the rooms to reach the green square randomly located in the bottom right room. 
Each MiniGrid environment is modified from the original implementation ~\cite{minigrid}. This modification demonstrates the ability of our method to generalize to new environments not seen during LLM pretraining.

\subsection{Baselines}
\label{sec:baselines}

In our experiments, we evaluate our method against several diverse baselines to comprehensively assess its performance in partially observable environments. One baseline, termed the \textit{oracle}, uses POMDP models that are hardcoded to exactly match the true dynamics of the environment, thereby serving as an upper-bound on achievable performance. In contrast, the \textit{random} baseline takes actions arbitrarily at every step, establishing a lower-bound benchmark for comparison.

Another baseline, referred to as \textit{direct LLM}, involves querying a large language model for the next action at each decision point. 
In this setup, the LLM is provided with all the same information provided to \Ours during model learning.
The exact prompt template used for this method is detailed in Appendix~\ref{app:direct_llm_prompt}. 
Next, our evaluation includes a \textit{behavior cloning} baseline, where a policy is constructed by mapping states to actions using a dictionary learned from the demonstration dataset. 
In addition, we consider a \textit{tabular baseline} in which the POMDP models are learned as conditional probability tables derived from counts in the demonstration dataset. 

Lastly, we test against two ablations of \OursNoSpace. The first is the \textit{offline only} ablation which only makes use of the human demonstrations and does not update the model with its own experiences. Conversely, the \textit{online only} ablation does not make use of the expert demonstrations, learning only from its own experiences. 

\subsection{Simulation Results}
\begin{figure*}[t!]
    \centering
    \includegraphics[width=1.0\textwidth]{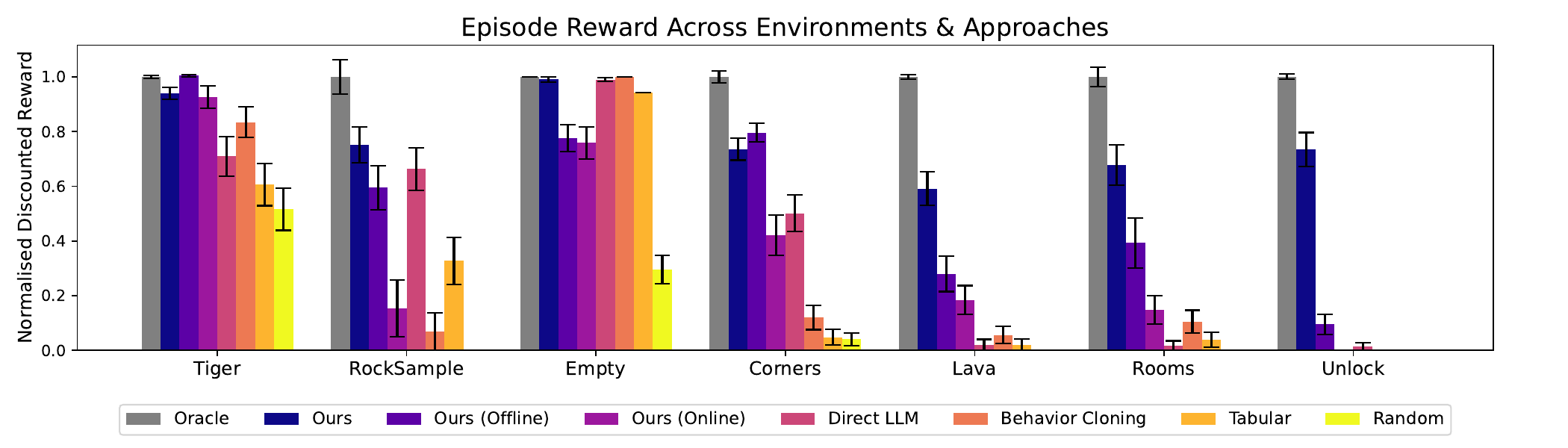}
    \caption{Experimental results for the MiniGrid and Classical POMDP domains. We show the expected discounted returns ($\gamma=0.98$) of each method across five learning seeds with ten episodes per seed. The error bars show standard error across all episodes. We normalize the expected discounted returns by the performance of Oracle.}
    \label{fig:results}
\end{figure*}

We evaluate various methods using expected discounted reward defined in Section~\ref{sec:background_pomdp}, measuring both total cumulative reward and efficiency.
The results of our evaluations can be seen in Figure~\ref{fig:results}. We see \Ours match or outperform all baseline methods across all domains. 

We observe that the behavior cloning and tabular baselines were fundamentally limited in their ability to generalize. This is because the set of possible initial states for many tasks was orders of magnitude larger than the training set. 
In contrast, the probabilistic programs written by \Ours use symbolic abstraction to cover large portions of the state and observation spaces, allowing them to generalize to new situations.
While the direct LLM approach was sometimes effective, such as in the rock sample domain, it frequently got stuck in infinite loops, failing to understand constraints such as obstacle obstruction despite the examples of collision it had access to in the dataset.

\Ours outperformed both the online-only and offline-only ablations across most environments. A common failure mode of the offline-only ablation was missing transitions outside expert demonstrations. For example, it will run into lava without knowing it causes death. In contrast, the online-only ablation struggled to discover informative actions due to inefficient random exploration. For instance, in the unlock environment, if the agent never used the key, the model failed to learn that behavior, and random actions rarely uncovered it.

In addition to success metrics, we record some additional runtime statistics such as the number of candidate programs generated during offline and online learning in Table~\ref{tab:nodes} as well as the training and testing coverage scores after offline model learning in Table~\ref{tab:coverages}.

\subsection{Real Robot Experiments}
\label{sec:real_robot_experiments}

Our real robot experiments are conducted using a Boston Dynamics Spot robot. The robot carries an in-hand camera mounted on a 6-DoF arm at its back. April tags are distributed throughout the area, enabling precise localization. The goal is for the robot to find an pick up an apple placed within the scene. Before any demonstrations are gathered, we construct a map of the empty room by scanning it with PolyCam~\cite{polycam}. This scan is used to construct a scene representation that includes an object-centric scene graph, encoding ``on'' relationships derived from geometric cues, and an occupancy grid delineating forbidden zones corresponding to physical obstacles. For each real-world task, ten demonstrations are collected by commanding the robot via keyboard.

The agent's action space is discretized into fixed theta rotations to the left and right and movements in the four cardinal directions.
Objects are detected online using Grounding SAM, an open vocabulary object detector~\cite{gsam1,gsam2,gsam3}.

\begin{figure*}[t!]
    \centering
    \includegraphics[width=1.0\textwidth]{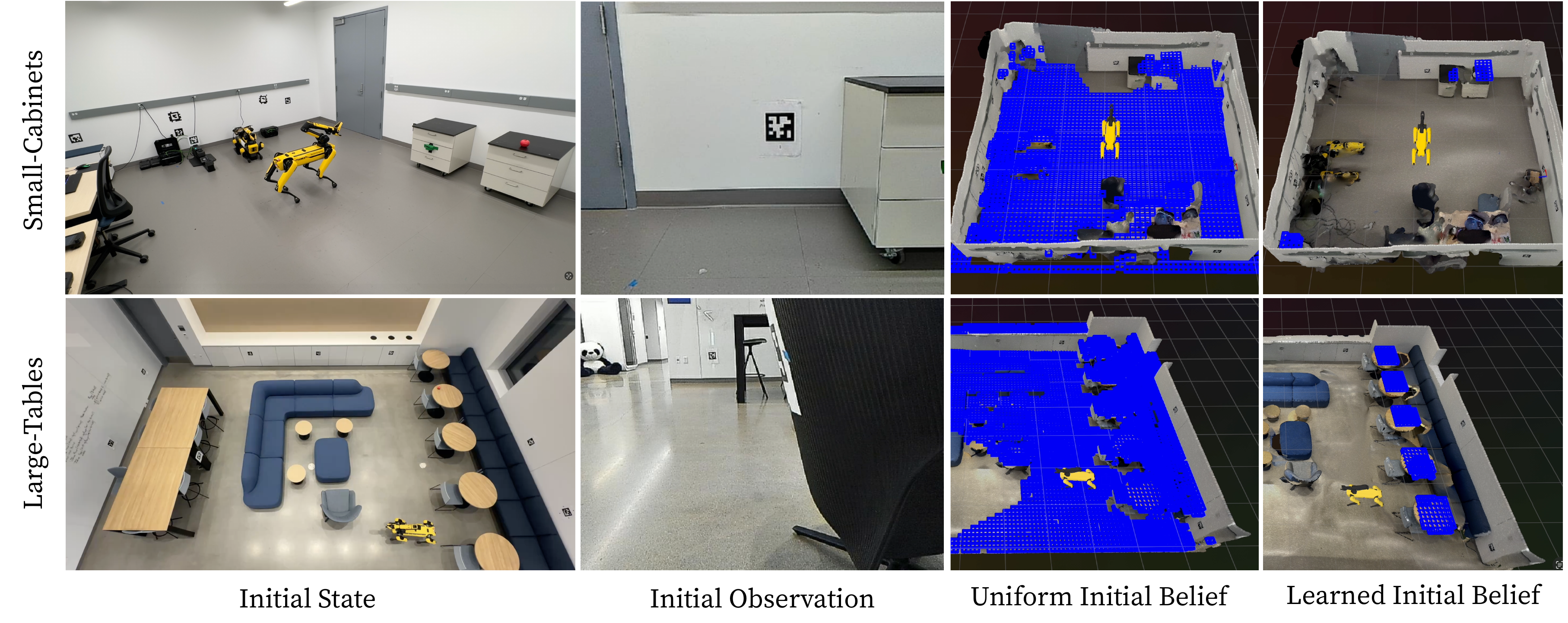}
    \caption{The two real-world experimental setups wherein a robot is searching for an apple in a partially observable world. The blue cells represent the robot's belief about where the apple could be in the world. In the uniform initial belief, the robot thinks the apple could be anywhere it has not looked yet. The learned initial belief found by \Ours has a narrower initial belief leading to more efficient exploration.}
    \label{fig:real_world_open}
\end{figure*}

We test our method in two distinct spaces. The first is a small, closed-off room, as shown in the top row of Figure~\ref{fig:real_world_open}, which contains a few tables, chairs, and drawer cabinets. 
The second is a large, open lobby area depicted in the bottom row of Figure~\ref{fig:real_world_open}, furnished with more than twenty pieces of furniture. 
Within these environments, task distributions are defined by varying the location of the apple. 
In the Small-Cabinets configuration, which takes place in the small room, the apple is consistently placed on top of one of the three drawer cabinets for each demonstration. 
In the Large-Tables setting within the large room, the apple is positioned on one of the five round tables.

Our approach is compared against a subset of baselines evaluated in Section~\ref{sec:baselines}. The evaluation includes behavior cloning, direct LLM execution, and a hardcoded uniform baseline that assumes the object is placed uniformly throughout the search space. 
Although the uniform baseline requires additional task-specific human input and is not strictly an apples-to-apples comparison, it demonstrates that our method can outperform a naively designed initial state distribution.

The results shown in Table~\ref{tab:real_world_results} demonstrate that \Ours achieves more efficient and accurate exploration by understanding and generalizing trends in initial state distribution seen in the training data. 
Specifically, our approach learns that objects are always on top of objects of a particular class, and constructs an initial state distribution that captures that without overfitting to specific initial states.

\begin{table}[ht]
\centering
\resizebox{\textwidth}{!}{%
\begin{tabular}{lccccc}
\toprule
               & Ours            & Uniform         & Direct LLM      & BC              & Tabular \\
\midrule
Small-Cabinets & $\textbf{0.89}\pm{\textbf{0.09}}\;(10)$ & $0.68\pm{0.21}\;(10)$ & $0.25\pm{0.42}\;(3)$ & $0.28\pm{0.39}\;(4)$ & $0.35\pm{0.41}\;(5)$ \\
Large-Tables   & $\textbf{0.73}\pm{\textbf{0.08}}\;(10)$ & $0.40\pm{0.26}\;(8)$ & $0.15\pm{0.32}\;(2)$ & $0.13\pm{0.28}\;(2)$ & $0.13\pm{0.22}\;(3)$ \\
\bottomrule
\end{tabular}}

\vspace{0.5em} 
\caption{Real-world experiment results on the small room domain in the top row of Figure~\ref{fig:real_world_open} and the large room domain in the bottom row of Figure~\ref{fig:real_world_open}. The table shows the mean and standard deviation of expected discounted reward (with $\gamma=0.98$) under the ground-truth reward model (1 if the agent is holding the apple and 0 otherwise) along with the number of successes over ten runs in parenthesis.}
\label{tab:real_world_results}
\end{table}

\section{Discussion}

In this work, we have presented a novel approach to learning interpretable, low-complexity POMDP models by integrating LLM-guided probabilistic program induction with online planning. Our method leverages large language models to generate and iteratively refine candidate probabilistic programs that capture the dynamics, observation functions, initial state distributions, and reward structures of complex environments. Experimental results on simulated MiniGrid domains and real-world robotics scenarios demonstrate that our approach can significantly enhance sample efficiency and predictive accuracy compared to traditional tabular learning methods, behavior cloning, or direct LLM planning.

Our findings further suggest that environments represented with structured scene graphs and other rich input representations can be better modeled by learning a world model within which a reasoning agent can operate, rather than by directly learning a policy that maps observation histories to actions or by attempting to apply a language model in a zero-shot setting. This is particularly evident in large, partially observable worlds where the belief space is considerably more complex and challenging to cover with training examples than the space of states itself. The use of code to represent these models is especially advantageous, as language models are adept at generating concise, executable snippets that can be interpreted, debugged, and evaluated post-hoc, thus providing an additional layer of transparency and robustness in model evaluation.

\newpage
\section{Limitations}
\label{sec:limitations}
Despite these promising results, several limitations remain. Our approach currently relies on human expertise to design the underlying representation over which the world model is learned, which may constrain its applicability to domains where such structured representations are not readily available. 

Additionally, due to our post-hoc observability assumption, collecting datasets outside of a simulator requires one of the following: human state annotation, complete robot exploration after the episode, or third-party perspectives such as externally mounted cameras.
In our real robot experiments, this was not a challenge because the only state variability was in the position of the goal object, which is fully determined upon completion of the task. 
More complex problems with multiple dimensions of both task-relevant and task-irrelevant uncertainty would require more than just the agent's perspective.
Alleviating this assumption may require jointly reasoning about the interrelated structure of the constituent models, and is a valuable direction for future work.

Moreover, the particle filter employed in our current implementation does not scale well to arbitrarily large state spaces. Future work may address this limitation by incorporating more advanced inference techniques, such as factored particle filters or other scalable methods, to improve performance in high-dimensional settings. 

Another area for improvement is in the sometimes overly broad distributions proposed by the LLM due to the coverage metric indirectly rewarding broader distributions. While this doesn't make the problem infeasible, it can lead to less efficient behavior. A direction for future work could be to use inference methods on the hidden variables of the proposed probabilistic program to strike a balance between the empirical distribution and the overly broad model distribution.

Lastly, our study focuses exclusively on discrete state and action spaces, despite robotics tasks requiring search over continuous spaces such as grasps and poses. Extending our learning strategy and adopting continuous-space POMDP solvers would broaden our framework to these domains, enabling more complex manipulation and navigation tasks.

Ultimately, our work opens up exciting avenues for combining the strengths of probabilistic programming and large language models to construct robust, interpretable models for decision-making under uncertainty.

\clearpage


\bibliography{example}  
\newpage
\appendix

\section{Code Release}
The full implementation, including all code, trained models, and experiment configurations, will be released publicly upon publication to ensure reproducibility and facilitate future research.

\section{Model Learning}
\label{app:model_learning}

The learning algorithm follows the pseudocode in Algorithm~\ref{alg:llmguidedlearning}. 
Firstly, in order to mitigate overfitting to spurious patterns, we split the demonstration dataset into separate training and test sets (Line~\ref{line:split}).
Given the training set and a code-based interface that specifies how states, actions, and observations are represented, we query an LLM to propose an initial code snippet for a given component (Line~\ref{line:initial_prompt}, see Appendix~\ref{app:initial_prompt} for prompt). 

\begin{algorithm}[ht]
\begin{spacing}{1.2} 
\begin{algorithmic}[1]
\State \textbf{Input:} demonstration dataset $\mathcal{D}$, initial model $\theta_{\text{prev}}$, budget $N$

\State \textbf{Output:} Learned model $\theta_{\text{new}}$

\State Split $\mathcal{D}$ into $\mathcal{D}_{\text{train}}, \mathcal{D}_{\text{test}}$ \label{line:split}

\State \textbf{if} $\theta=\emptyset$ \textbf{then} $\theta_{\text{prev}} \gets \text{LLM Init using } \mathcal{D}_{\text{train}}$ \Comment{Appendix~\ref{app:initial_prompt}} \label{line:initial_prompt}

\State $\text{coverage}, \mathcal{D}_{\text{errors}} \gets \texttt{Eval}(\theta_{\text{prev}}, \mathcal{D}_{\text{test}}, \mathcal{D}_{\text{train}})$ \Comment{Discrepancy between model \& empirical}\label{line:eval}

\State $\text{beta} \gets \text{Beta}(1+C\cdot\text{coverage}, 1+C\cdot(1-\text{coverage}))$
\State $\mathcal{M} \gets \{(\theta_{\text{prev}}, \text{beta}, \mathcal{D}_{\text{errors}})\}$ \label{line:root}
\While{$\text{coverage} < 1.0$ \textbf{and} $\text{iterations} < M$}
    \State $(\theta_{\text{new}}, \text{Beta}(\alpha, \beta), \mathcal{D}_{\text{errors}}) \gets \text{argmax}_\mathcal{M}(p\sim\text{beta})$ \Comment{Thompson sampling} \label{line:thompson}
    \State $\theta_{\text{new}}^\prime \gets$ LLM refinement using $\theta_{\text{new}}$ and $\mathcal{D}_{\text{errors}}$ \Comment{Appendix~\ref{app:refinement_prompt}} \label{line:refinement}
    \State $\text{coverage}^\prime, \mathcal{D}_{\text{errors}}^\prime \gets \texttt{Eval}(\theta_{\text{new}}, \mathcal{D}_{\text{test}}, \mathcal{D}_{\text{train}})$ \label{line:eval2}
    \State $\text{beta}^\prime \gets \text{Beta}(\alpha + C\cdot\text{coverage}^\prime, \beta + C\cdot(1-\text{coverage}^\prime))$
    \State \textbf{insert} $(\theta_{\text{new}}^\prime, \text{beta}^\prime, \text{coverage}^\prime)$ \textbf{into} $\mathcal{M}$ \label{line:insert}
\EndWhile
\State \Return $\text{argmax}_\mathcal{M}(\text{coverage})[0]$  \Comment{Return the model with the best overall coverage} \label{line:return}
\end{algorithmic}
\end{spacing}
\caption{\texttt{LearnModel}}
\label{alg:llmguidedlearning}
\end{algorithm}

Following the initial candidate model proposal, we evaluate them against the empirical conditional probability distributions observed in the demonstration data (Line~\ref{line:eval}).
Determining if a particular outcome is possible under an arbitrary code model is not analytically possible in most probabilistic programming languages, including Pyro, so we use Monte Carlo approximation of the model density. 
Any empirical sample that is never produced is treated as a failure and recorded in an error set \(\mathcal{D}_\text{errors}\).
During evaluation, we estimate the model's coverage on a combination of the training and testing sets, evaluating models based on their ability to generalize beyond the training examples (Line~\ref{line:eval}). 

We proceed with an iterative learning procedure that uses a Thompson sampling exploration strategy to build out a tree of candidate models. In addition to a candidate code block, each node contains a Beta distribution capturing uncertainty over its true coverage performance. Initially, the root node contains the LLM's first code proposal evaluated against the data (Line~\ref{line:root}). At each iteration, we select a node to expand using Thompson sampling: we sample from each node's Beta distribution and pick the node with the highest sampled value (Line~\ref{line:thompson}).

The selected node is refined by prompting the LLM with its associated training set coverage mismatch errors \(\mathcal{D}_{\text{errors}}\), encouraging the LLM to generate a corrected or improved version of the model (Line~\ref{line:refinement}, see Appendix~\ref{app:refinement_prompt} for prompt). This produces a child node with updated code, which is re-evaluated to obtain new train and test coverage statistics (Line~\ref{line:eval2}). The parent's Beta distribution is updated using a smoothing constant \(C\) to encourage stable learning from finite samples, and the child node is added to the tree (Line~\ref{line:insert}).

This iterative process continues until the overall coverage across the empirical distribution is sufficiently high, or until a pre-defined iteration budget is exhausted. Ultimately, we return the candidate model with the highest empirical coverage among all nodes (Line~\ref{line:return}).

\section{Belief-Space Planner}
\label{app:planner}

\begin{algorithm}[ht]
\begin{spacing}{1.2}
\begin{algorithmic}[1]
\State \textbf{Input:} initial belief $b_0$, models $\mathcal{T}$, $\mathcal{O}$, $\mathcal{R}$, horizon $H$, hyper-parameters $\lambda,\alpha$
\State \textbf{Output:} best first action $a^\ast$
\vspace{2pt}
\State $\text{Open} \gets \{(b_0, g{=}0)\}$, \text{Closed}$\gets\emptyset$, \text{Cost}$\gets\{b_0: 0\}$ \label{line:init}
\While{\text{Open} $\neq \emptyset$ \textbf{and} iterations $< H$} \label{line:while}
    \State $(b, g) \gets \text{pop\_lowest}(\text{Open})$
    \If{$b$ is terminal} \label{line:terminalcheck}
        \State \textbf{continue}
    \EndIf
    \If{$b \in$ \text{Closed}} \label{line:closedcheck}
        \State \textbf{continue}
    \EndIf
    \State add $b$ to \text{Closed}
    \For{\textbf{each} action $a \in \mathcal{A}$} \label{line:actionloop}
        \State Draw $n$ state particles $s' \sim \mathcal{T}(s,a)$ for $s \sim b$ \label{line:sampleT}
        \State Draw observations $o \sim \mathcal{O}(s',a)$ \label{line:sampleO}
        \State Form child belief $b' = \texttt{Branch}(b,a,o)$ \label{line:branch}
        \State $\hat r \gets \mathbb{E}_{s,s'}[\mathcal{R}(s,a,s')]$ \label{line:reward}
        \State $\hat p \gets \Pr[o\mid b,a]$, \quad $\hat h \gets H(b')$ \label{line:probentropy}
        \State $g' \gets g - \hat r - \lambda \log \hat p + \alpha \hat h + \text{cost}(a)$ \label{line:gprime}
        \If{$b' \notin$ \text{Cost} \textbf{or} $g' < \text{Cost}[b']$} \label{line:insertopen}
            \State $\text{Cost}[b'] \gets g'$
            \State insert $(b', g')$ into \text{Open}
        \EndIf
    \EndFor
\EndWhile
\State \Return first action in the path to the node in \text{Open }$\cup$\text{ Closed} with minimal $g$ \label{line:return}
\end{algorithmic}
\end{spacing}
\caption{\texttt{BeliefPlanner}}
\label{alg:beliefplanner}
\end{algorithm}

Our online planning routine conducts forward search directly in belief space, determinizing the stochastic dynamics to enable an A*-style expansion strategy that balances exploitation (reward) and exploration (information gain). 
Algorithm~\ref{alg:beliefplanner} provides the complete procedure.

We begin by inserting the initial belief \(b_0\) into an \emph{open} priority queue with zero cost-to-come and initializing an empty \emph{closed} set (Line~\ref{line:init}). 
Each queue element stores the belief, its cumulative cost \(g\), and bookkeeping metadata such as depth. 
During each iteration (Line~\ref{line:while}), we pop the node with the lowest priority value; if it is terminal or has already been expanded (i.e., in the closed set), we skip further expansion (Lines~\ref{line:terminalcheck}–\ref{line:closedcheck}).

Otherwise, for every action \(a\) (Line~\ref{line:actionloop}), we draw next-state particles from the transition model \(\mathcal{T}\) (Line~\ref{line:sampleT}) and sample the corresponding observations through the observation model \(\mathcal{O}\) (Line~\ref{line:sampleO}).  
Conditioning on the sampled observation yields a child belief \(b'\) (Line~\ref{line:branch}).  
We estimate the expected reward \(\hat r\) under \(\mathcal{R}\), the likelihood \(\hat p\) of the observation, and the entropy \(\hat h\) of \(b'\) (Lines~\ref{line:reward}–\ref{line:probentropy}).  
These statistics define the child’s cost 
\[
g' \;=\; g \;-\; \hat r \;-\; \lambda \log \hat p \;+\; \alpha \hat h \;+\; \text{cost}(a),
\]
where \(\lambda\) and \(\alpha\) trade off risk sensitivity and information gain (Line~\ref{line:gprime}).  
The child node is inserted into the queue only if it is not yet discovered or has a lower cumulative cost than a previously seen version (Line~\ref{line:insertopen}). 
The process continues until the queue is empty or a computational budget is exhausted.

Finally, we return the first action in the path from \(b_0\) to the node with the minimum accumulated cost (Line~\ref{line:return}), thereby maximizing the composite objective of long-term reward, low risk, and maximal information gathering.

While this planner is similar in many ways to POUCT~\cite{POMCPOW}, it has the additional feature that enables graph-based search rather than strictly tree-based search, which proved computationally necessary for many of our larger MiniGrid problems. It is important to note that this planner is not optimal, but it suffices for all of the problems we tested, and can easily be substituted for other planning methods in the \Ours framework.

\section{Hyperparameters}
Table~\ref{tab:hyperparameters} shows the hyperparameters used per domain for both planning and learning. The planning hyperparameters (in grey) were tuned to work best for the ground truth models used in oracle. 
With the exception of Thompson smoothing coefficient which was selected based on ~\cite{rex}, the other hyperparameters were selected to be as large as possible under computational and budgetary constraints.

\begin{table}[h]
\centering
\begin{tabular}{lccc}
\toprule
\textbf{Hyperparameter} & \textbf{Classical} & \textbf{MiniGrid} & \textbf{Spot Robot} \\
\midrule
\rowcolor{gray!10}
Action cost penalty                      & 0.01  & 0.01  & 0.01  \\
\rowcolor{gray!10}
$\alpha$ (Entropy coefficient)           & 0.0   & 0.0   & 1.0   \\
\rowcolor{gray!10}
$\lambda$ (log-prob reward shaping)      & 0.1   & 0.1   & 0.1   \\
\rowcolor{gray!10}
Rollouts per stochastic model query      & 5     & 1     & 1     \\
\rowcolor{gray!10}
$H$ (Max expansions)                     & 50    & 5000  & 5000  \\
$N$ (Num initial particles),             & 50    & 10    & 10    \\
Max particle rejuvenations,            & 2,500,00 & 500,000 & 25,000 \\
$M$ Max refinements  & 25    & 25    & 25    \\ 
$C$ Thompson smoothing & 25    & 25    & 25    \\
$ND$ (\# datapoints shown - initial, \ref{app:initial_prompt})     & 5    & 5    & 5    \\
$NC$ (\# conditions shown - refinement, \ref{app:refinement_prompt})     & 5  & 5   & 5    \\
$NS$ (\# samples per condition - refinement, \ref{app:refinement_prompt}) & 5   & 5   & 5  \\
\bottomrule
\end{tabular}
\label{tab:hyperparameters}
\end{table}

\section{Prompts}

\subsection{Function Templates}

\begin{minted}[fontsize=\small, breaklines, tabsize=4, bgcolor=LightGray]{python}

def initial_func(empty_state:MiniGridState):
    """
    Input:
        empty_state (MiniGridState): An empty state with only the walls filled into the grid
    Returns:
        state (MiniGridState): the initial state of the environment
    """
    raise NotImplementedError
    
\end{minted}

\begin{minted}[fontsize=\small, breaklines, tabsize=4, bgcolor=LightGray]{python}

def observation_func(state, action, empty_obs):
    """
    Args:
        state (MiniGridState): the state of the environment
        action (int): the previous action that was taken
        empty_obs (MiniGridObservation): an empty observation that needs to be filled and returned
    Returns:
        obs (MiniGridObservation): observation of the agent
    """
    raise NotImplementedError
    
\end{minted}

\begin{minted}[fontsize=\small, breaklines, tabsize=4, bgcolor=LightGray]{python}

def reward_func(state, action, next_state):
    """
    Args:
        state (MiniGridState): the state of the environment
        action (int): the action to be executed
        next_state (MiniGridState): the next state of the environment
    Returns:
        reward (float): the reward of that state
        done (bool): whether the episode is done
    """
    raise NotImplementedError
    
\end{minted}

\begin{minted}[fontsize=\small, breaklines, tabsize=4, bgcolor=LightGray]{python}

def transition_func(state, action):
    """
    Args:
        state (MiniGridState): the state of the environment 
        action (int): action to be taken in state `state`
    Returns:
        new_state (MiniGridState): the new state of the environment
    """
    raise NotImplementedError
    
\end{minted}

\subsection{Initial Prompt}
\label{app:initial_prompt}

\begin{minted}[fontsize=\small, breaklines, tabsize=4, bgcolor=LightGray]{text}
You are a robot exploring its environment. 

Environment Description: {env_description}
Goal Description: {goal_description}

Your goal is to model the {what_to_model}. 
You need to implement the python code to model the world, as seen in the provided experiences. 
Please follow the template to implement the code. 
The code needs to be directly runnable {model_input} and return {model_output}. 


Below are a few samples from the environment distribution. These are only samples from a larger distribution that your should model.

{exp}

Here is the template for the {model_name} function. Please implement
the reward function following the template. The code needs to be directly
runnable.

‘‘‘
{code_api}

{code_template}
‘‘‘

Explain what you believe is the {what_to_model} in english.
Additionally, please implement code to model the logic of the world. Please implement the 
code following the template. Only output the definition for ‘ {model_name} ‘. 
You must implement the ‘ {model_name} ‘ function.
Create any helper function inside the scope of ‘ {model_name} ‘. 
Do not create any helper function outside the scope of ‘ {model_name} ‘.
Do not output examples usage. 
Do not create any new classes.
Do not rewrite existing classes. 
Do not import any new modules from anywhere.
Do not overfit to the specific samples.
Put the ‘ {model_name} ‘ function in a python code block.
Implement any randomness with `pyro.sample`
    
\end{minted}

\subsection{Refinement Prompt}

\label{app:refinement_prompt}
\begin{minted}[fontsize=\small, breaklines, tabsize=4, bgcolor=LightGray]{text}
You are a robot exploring its environment. 

{env_description}

Your goal is to model {what_to_model} of the world in python. 

You have tried it before and came up with one partially correct solution, but it is not perfect. 

The observed distribution disagrees with the generated model in several cases.
You need to improve your code to come closer to the true distribution.

Environment Description: {env_description}
Goal Description: {goal_description}

Here is a solution you came up with before. 

```
{code_api}

{code}
```


{experiences}

Explain what you believe is {what_to_model} in english, then improve your code to better model the true distribution.

Please implement the code for the following the template. 
You must implement the ‘ {model_name} ‘ function. 

The code needs to be directly runnable {model_input} and return {model_output}. 

Do not output examples. 
Do not create any new classes. 
Do not rewrite existing classes.
Do not import any new modules from anywhere.
Do not list out specific indices that overfit to the examples, but include ranges.
Put the ‘ {model_name} ‘ function in a python code block.
Implement any randomness with `pyro.sample`
\end{minted}

The experiences in the refinement prompt are structured as follows. First, we sample conditions for which there is coverage less than 1. For example, in the transition model $P(s_{t+1}|s_{t},a_{t})$, we search through the set of $(s_{t}, a_{t})$ tuples in the database and we find the set of those tuples where the distribution over $(s_{t+1})$ contains at least 1 element that can not be achieved by the LLM-generated model. We select out $NC$ of those conditions. Then, given those conditions, we select $NS$ samples that were not covered by the model to show as examples to the LLM. An example template for a single condition and an $NS=3$ is shown below. The values we used for $NC$, $NS$ are in Table~\ref{tab:hyperparameters}.

\begin{minted}[fontsize=\small, breaklines, tabsize=4, bgcolor=LightGray]{text}
Here are some samples from the real world that were impossible under your model
{condition} -> {dataset_outcome}
{condition} -> {dataset_outcome}
{condition} -> {dataset_outcome}

And here are some samples from your code under the same conditions
{condition} -> {model_outcome}
{condition} -> {model_outcome}
{condition} -> {model_outcome}

\end{minted}

\subsection{Direct LLM Baseline Prompt}
\label{app:direct_llm_prompt}

\begin{minted}[fontsize=\small, breaklines, tabsize=4, bgcolor=LightGray]{text}
You are a robot exploring its environment. 

{env_description}

Your goal is to predict the next best action to take to reach the goal and maximize reward.

Here is the template for the reward function. Please implement
the reward function following the template. The code needs to be directly
runnable on the inputs of (state) and return (reward) in python.

‘‘‘
{code_api}
‘‘‘

Here are some example rollouts from the environment

{exp}

Here is the current episode history for the task that you are doing right now

{current_episode}

Output the next aciton in the form where you fill in <action-here> with the action that is best for reaching the goal and maximizing reward.
For example, your code will look like this:

```
next_action:int = 0
```

The action should be an integer with no additional code. Explan your reasoning in one sentence.

\end{minted}

\subsection{Runtime Statistics}

\begin{table}[h]
\resizebox{\textwidth}{!}{%
\begin{tabular}{lccccccc}
\toprule
Approach & Tiger & RockSample & Empty & Corners & Lava & Rooms & Unlock \\
\midrule
Offline Transition & $2.00\,\pm\,0.00$ & $2.60\,\pm\,0.24$ & $2.20\,\pm\,0.20$ & $2.00\,\pm\,0.00$ & $2.00\,\pm\,0.00$ & $2.60\,\pm\,0.24$ & $4.60\,\pm\,0.81$ \\
Online Transition & $0.00\,\pm\,0.00$ & $0.06\,\pm\,0.06$ & $0.00\,\pm\,0.00$ & $0.00\,\pm\,0.00$ & $0.00\,\pm\,0.00$ & $0.00\,\pm\,0.00$ & $0.40\,\pm\,0.15$ \\
Offline Reward & $2.40\,\pm\,0.24$ & $11.80\,\pm\,4.33$ & $2.00\,\pm\,0.00$ & $2.00\,\pm\,0.00$ & $2.20\,\pm\,0.20$ & $2.00\,\pm\,0.00$ & $2.20\,\pm\,0.20$ \\
Online Reward & $0.05\,\pm\,0.05$ & $0.05\,\pm\,0.05$ & $0.00\,\pm\,0.00$ & $0.00\,\pm\,0.00$ & $0.19\,\pm\,0.09$ & $0.00\,\pm\,0.00$ & $0.00\,\pm\,0.00$ \\
Offline Observation & $2.20\,\pm\,0.20$ & $5.80\,\pm\,1.46$ & $10.40\,\pm\,4.01$ & $7.80\,\pm\,4.61$ & $2.40\,\pm\,0.24$ & $17.25\,\pm\,5.71$ & $15.00\,\pm\,3.63$ \\
Online Observation & $0.00\,\pm\,0.00$ & $0.00\,\pm\,0.00$ & $0.05\,\pm\,0.05$ & $0.08\,\pm\,0.06$ & $0.00\,\pm\,0.00$ & $0.28\,\pm\,0.17$ & $0.05\,\pm\,0.05$ \\
Offline Initial & $2.00\,\pm\,0.00$ & $2.60\,\pm\,0.24$ & $2.00\,\pm\,0.00$ & $2.80\,\pm\,0.37$ & $22.60\,\pm\,3.40$ & $1.75\,\pm\,0.25$ & $18.80\,\pm\,4.59$ \\
Online Initial & $0.00\,\pm\,0.00$ & $0.02\,\pm\,0.02$ & $0.00\,\pm\,0.00$ & $0.17\,\pm\,0.09$ & $0.86\,\pm\,0.21$ & $0.06\,\pm\,0.06$ & $0.49\,\pm\,0.15$ \\
\bottomrule
\end{tabular}
}
\vspace{0.5em}
\caption{The average and standard deviation of the number of nodes, or LLM-generated candidate programs, sampled during the online and offline phases of model learning.}
\label{tab:nodes}
\end{table}

\begin{table}[h]
\resizebox{\textwidth}{!}{%
\begin{tabular}{lccccccc}
\toprule
Approach & Tiger & RockSample & Empty & Corners & Lava & Rooms & Unlock \\
\midrule
Transition Train & $1.00\,\pm\,0.00$ & $1.00\,\pm\,0.00$ & $1.00\,\pm\,0.00$ & $1.00\,\pm\,0.00$ & $1.00\,\pm\,0.00$ & $1.00\,\pm\,0.00$ & $1.00\,\pm\,0.00$ \\
Transition Test & $1.00\,\pm\,0.00$ & $1.00\,\pm\,0.00$ & $1.00\,\pm\,0.00$ & $1.00\,\pm\,0.00$ & $1.00\,\pm\,0.00$ & $1.00\,\pm\,0.00$ & $1.00\,\pm\,0.00$ \\
Reward Train & $1.00\,\pm\,0.00$ & $0.99\,\pm\,0.01$ & $1.00\,\pm\,0.00$ & $1.00\,\pm\,0.00$ & $1.00\,\pm\,0.00$ & $1.00\,\pm\,0.00$ & $1.00\,\pm\,0.00$ \\
Reward Test & $1.00\,\pm\,0.00$ & $0.99\,\pm\,0.01$ & $1.00\,\pm\,0.00$ & $1.00\,\pm\,0.00$ & $1.00\,\pm\,0.00$ & $1.00\,\pm\,0.00$ & $1.00\,\pm\,0.00$ \\
Observation Train & $1.00\,\pm\,0.00$ & $1.00\,\pm\,0.00$ & $0.92\,\pm\,0.08$ & $0.81\,\pm\,0.19$ & $1.00\,\pm\,0.00$ & $0.94\,\pm\,0.06$ & $0.94\,\pm\,0.06$ \\
Observation Test & $1.00\,\pm\,0.00$ & $1.00\,\pm\,0.00$ & $0.92\,\pm\,0.08$ & $0.88\,\pm\,0.12$ & $1.00\,\pm\,0.00$ & $0.93\,\pm\,0.06$ & $0.94\,\pm\,0.06$ \\
Initial Train & $1.00\,\pm\,0.00$ & $1.00\,\pm\,0.00$ & $1.00\,\pm\,0.00$ & $1.00\,\pm\,0.00$ & $0.68\,\pm\,0.16$ & $0.75\,\pm\,0.25$ & $0.88\,\pm\,0.12$ \\
Initial Test & $1.00\,\pm\,0.00$ & $1.00\,\pm\,0.00$ & $1.00\,\pm\,0.00$ & $1.00\,\pm\,0.00$ & $0.56\,\pm\,0.18$ & $0.75\,\pm\,0.25$ & $0.24\,\pm\,0.19$ \\
\bottomrule
\end{tabular}
}
\vspace{0.5em}
\caption{The average and standard deviation of coverages achieved after the offline model learning step has completed running split into training and testing coverages.}
\label{tab:coverages}
\end{table}

\end{document}